\documentclass[10pt,twocolumn,letterpaper]{article}

\usepackage{iccv}
\usepackage{times}
\usepackage{epsfig}
\usepackage{graphicx}
\usepackage{amsmath}
\usepackage{amssymb}
\usepackage[caption=false]{subfig}
\usepackage{booktabs}
\usepackage[accsupp]{axessibility} 

\usepackage[pagebackref=true,breaklinks=true,letterpaper=true,colorlinks,bookmarks=false]{hyperref}

\iccvfinalcopy 

\def\iccvPaperID{****} 

\ificcvfinal\fi

\begin{document}

\title{Border-SegGCN: Improving Semantic Segmentation by Refining the Border Outline using Graph Convolutional Network}
\author{Naina Dhingra\\
ETH Zurich\\
{\tt\small ndhingra@ethz.ch}
\and
George Chogovadze\\
ETH Zurich\\
{\tt\small chogeorg@student.ethz.ch	}
\and
Andreas Kunz\\
ETH Zurich\\
{\tt\small kunz@iwf.mavt.ethz.ch}}

\maketitle
\ificcvfinal\thispagestyle{empty}\fi


\begin{abstract}
We present Border-SegGCN, a novel architecture to improve semantic segmentation by refining the border outline using graph convolutional networks (GCN). The semantic segmentation network such as Unet or DeepLabV3+ is used as a base network to have pre-segmented output. This output is converted into a graphical structure and fed into the GCN to improve the border pixel prediction of the pre-segmented output. We explored and studied the factors such as border thickness, number of edges for a node, and the number of features to be fed into the GCN by performing experiments. We demonstrate the effectiveness of the Border-SegGCN on the CamVid and Carla dataset, achieving a test set performance of 81.96\% without any post-processing on CamVid dataset. It is higher than the reported state of the art mIoU achieved on CamVid dataset by 0.404\%. 

\end{abstract}

\section{Introduction}
\textbf{Semantic segmentation} provides dense per pixel classification corresponding to  each pixel's labels \cite{caesar2018coco,cordts2016cityscapes,everingham2015pascal}, which is a crucial task in the field of computer vision. Solving and understanding the environment within an image using segmentation opens the door to many applications in the field of robotics, security, autonomous vehicles, helping  blind and visually impaired people (BVIP), etc. With the advancements in deep neural networks,  \cite{dhingra2019res3atn,rakelly2018conditional,krizhevsky2012imagenet,rota2018place}, computer vision tasks such as semantic segmentation performance have improved using large scale datasets for training as compared to segmentation using handcrafted features \cite{he2004multiscale,ladicky2009associative,yao2012describing,gould2009decomposing}. Segmentation of pixels in non-boundaries regions is easier than for the object boundaries, since border pixels are prone to have large ambiguity in belonging to a particular segment. 

\textbf{Graph convolutional network} 
Research in the field of Graph Neural Networks (GNN) have produced interesting results for graph classification that do not require as much resources as traditional Neural networks \cite{DBLP:journals/corr/KipfW16}. Graph Convolutional Networks (GCN) have been used in the field of body pose estimation, action recognition, etc, as human body data can be converted into graphical forms \cite{wang2020global,zheng2019fall,zhao2019semantic,yan2018spatial,li2019actional,gao2019optimized,liu2019si}. Similarly, point cloud data has also been used in graphical forms. \cite{qian2019pu,xu2019grid,li2019deepgcns}.

\begin{figure}[!htb]
    \centering
    \includegraphics[width=1\linewidth, height=5cm]{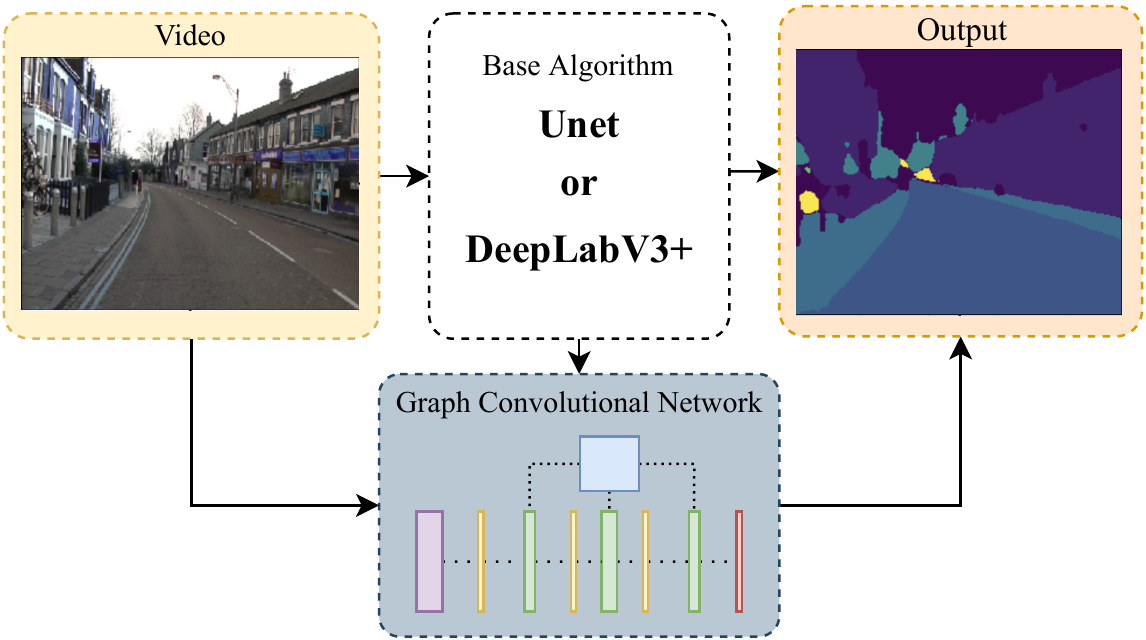} 
    \vspace{0.25cm}
     \caption[Simple overview]{We aim to adapt a semantic segmentation model learned from base algorithm using a graph convolutional network (GCN). The initial segmentation from base algorithm is used as basis and thereafter, GCNs are used to improve the object border outline pixel classification.}
    \label{fig:Simple overview}
\end{figure}

 As illustrated in Figure \ref{fig:Simple overview}, we propose a novel GCN based architecture for refining the outline of object boundaries on the pre-segmented image from a baseline network such as Unet or any other arbitrary network. Instead of focusing on the whole video segmentation task using GCNs, we focus on improving the object boundaries causing more ambiguities at the boundaries than at the non-border regions. The loss function penalizes for the border pixels in the network. It also decreases the computation complexity for the whole network as it has to only focus on some regions of the whole scene in each frame of a video. We effectively build a pipeline to generate graph based data from the visual data using spatial and intensity information of the pixels. Our architecture gives the flexibility to use any baseline network for the pre-segmentation task. We performed evaluation using two baseline networks to study the effect of using very different baseline network. In addition, a visualization analysis reveals that the object outline segmentation learned by the proposed architecture has meaningful semantic predictions.

Our contributions are summarized as follows: (1) We develop a novel method, Border-SegGCN, using GCNs to improve the semantic segmentation performance by refining the predicted boundaries. (2) We test our pipeline using two open source datasets, i.e., Camvid and Carla dataset (3) We perform the evaluation of different border widths to be used for a better performance of the procedure. (4) We evaluate the performance with a varying number of edges for the nodes in the graph. (5) We study the effect of a different number of features used as an input to the GCN network. (6) We evaluate the training with different hyper-parameters such as dropout and regularisation. (7) We also verify the performance of Border-SegGCN visually using qualitative results. (8) We prove the performance of the network using two different baseline models, i.e., Unet and DeepLabV3+. Border-SegGCN consistently improves performance when used on top of baseline approaches.

\section{Related Work}
\textbf{Semantic segmentation} is actively researched by computer vision researchers due to its application in various domains such as robotics, health care, security, etc. Most recent work is using fully convolutional neural networks for pixel-wise classification \cite{rakelly2018conditional,mou2019relation,zhang2018fully,li2017fully,lin2016scribblesup}. Various model variants have been researched \cite{chen2018encoder,schwing2015fully,jampani2016learning,vemulapalli2016gaussian} to use the contextual information for the task of segmentation \cite{mostajabi2015feedforward,dai2015convolutional,yao2012describing,ladicky2009associative}  based on multi-scale inputs \cite{farabet2012learning,eigen2015predicting,lin2016efficient} or based on probabilistic graphical models \cite{pinheiro2014recurrent,chandra2016fast,chandra2017dense,chen2017deeplab}. There are many open-source datasets available for semantic segmentation, such as e.g., PASCAL \cite{everingham2010pascal}, COCO \cite{lin2014microsoft}, CamVid \cite{CamVid}, Carla \cite{dosovitskiy2017carla}, etc. Our work uses the CamVid and Carla dataset for the semantic segmentation task. We use Unet and DeepLabV3+ as our base networks for the segmentation.

\textbf{Graph neural networks} have recently been used for various applications involving the graph structured data because of their effectiveness in representing such data \cite{xie2018memory}. Some of the tasks are action recognition, body pose estimation, link prediction, etc. Graph models can be subdivided into two categories, i.e, Graph Neural Network (GNN) \cite{xu2018powerful,wu2019comprehensive,qi20173d,monti2017geometric,zhang2018link,li2017situation,jaume2018image}, and GCNs. 

The first category, i.e., GNN, consists of a graph and a recurrent neural network and has a functionality of passing messages and updating nodes states which store the semantic and structural information in the neighboring nodes. For instance, \cite{qi20173d} uses a 3D graph neural network (3DGNN) with a 3D point cloud to build a k-nearest neighbor graph for RGBD semantic segmentation. Every node in the graph represents a group of points which has a hidden representation vector that is updated based on the recurrent functions.

The second category, i.e., GCN, extends the mathematical operation of convolution to graph structures. GCNs can be distinguished into two types, i.e., spectral GCNs and spatial GCNs. Spectral GCNs convert signals in graphical form into graph spectral domains, in which they can be manipulated using spectral filters. For instance, in \cite{duvenaud2015convolutional,henaff2015deep}, graph Laplacian-based CNNs are used in the spectral domain. Spectral GCNs use convolution along with neighborhood data to calculate the feature vector for the node. They were first employed for semi-supervised classification on graph-structured data \cite{DBLP:journals/corr/KipfW16}. For example,  \cite{simonovsky2017dynamic} use GCNs on point cloud data in the spatial domain. \cite{si2019attention} use  GCNs in a recurrent structure for skeleton based action recognition.

Both \cite{graphcut} and \cite{zhang15multiclass} have shown that image semantic segmentation can be formulated and solved using graphs. This paper focuses on applying GCNs in the context of video segmentation. The main issues discussed revolve around creating an efficient pipeline to convert video frames into graphs and optimising the pipeline to maximizing the efficiency of the GCN.

\section{Methodology}
\label{chp:Our approach}
Figure \ref{fig:Simple overview} illustrates an overview of the used pipeline. In the following section, we will discuss how we approach the graph creation from an image.
\subsection{Efficient Graph Generation}
\label{sec:Efficient graph generation}

This approach includes a refinement of borders using existing segmentation algorithms. We use the Unet \cite{ronneberger2015u} and DeepLabv3+ architecture as a base segmentation networks. The dataset used for the evaluation of this architecture is the Cambridge-driving Labeled Video Database ("CamVid") dataset \cite{BrostowFC:PRL2008} and Carla dataset \cite{dosovitskiy2017carla}. The frame size is 360 pixels wide and 480 pixels high. A mask is used to determine which nodes to compute based on their location, while omitting the rest of the nodes. This allows the graph to have the same size from frame to frame. Performing computation on the selected nodes allows decreasing computation time drastically and provides more relevant training examples to the GCN.

\begin{figure}[!htb]
    \centering
    \includegraphics[width=\linewidth, height=2.2cm]{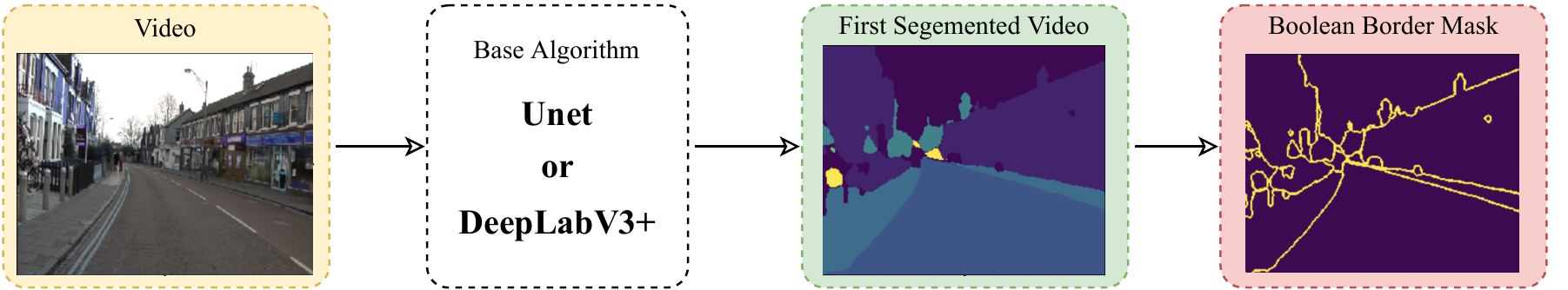}
        \vspace{0.25cm}
    \caption[Boolean mask generation example]{Example of how the data is processed during the Boolean mask generation. 
    Left: The initial frame is passed on to the base segmentation algorithm. Center: An initial rough segmentation is retrieved. Right: Pixels are either selected (yellow) or rejected (purple) depending if they are on the border of different objects.}
    \label{fig:Boolean mask generation example}
\end{figure}

To determine which nodes are selected we use the initial segmentation that our base algorithm provides: in this case the Unet or DeepLabV3+. With the pre-segmentation provided, boundaries around the objects and different classes are determined. Depending on the task at hand, the pixels that lay on the border are then selected for further processing followed by training or prediction. Figure \ref{fig:Boolean mask generation example} shows the process of generating the Boolean mask.

\subsection{Generating the Features}
\label{subsec:Generating the features}

\begin{figure}[!htb]
    \centering
    \includegraphics[width=\linewidth, height=6cm]{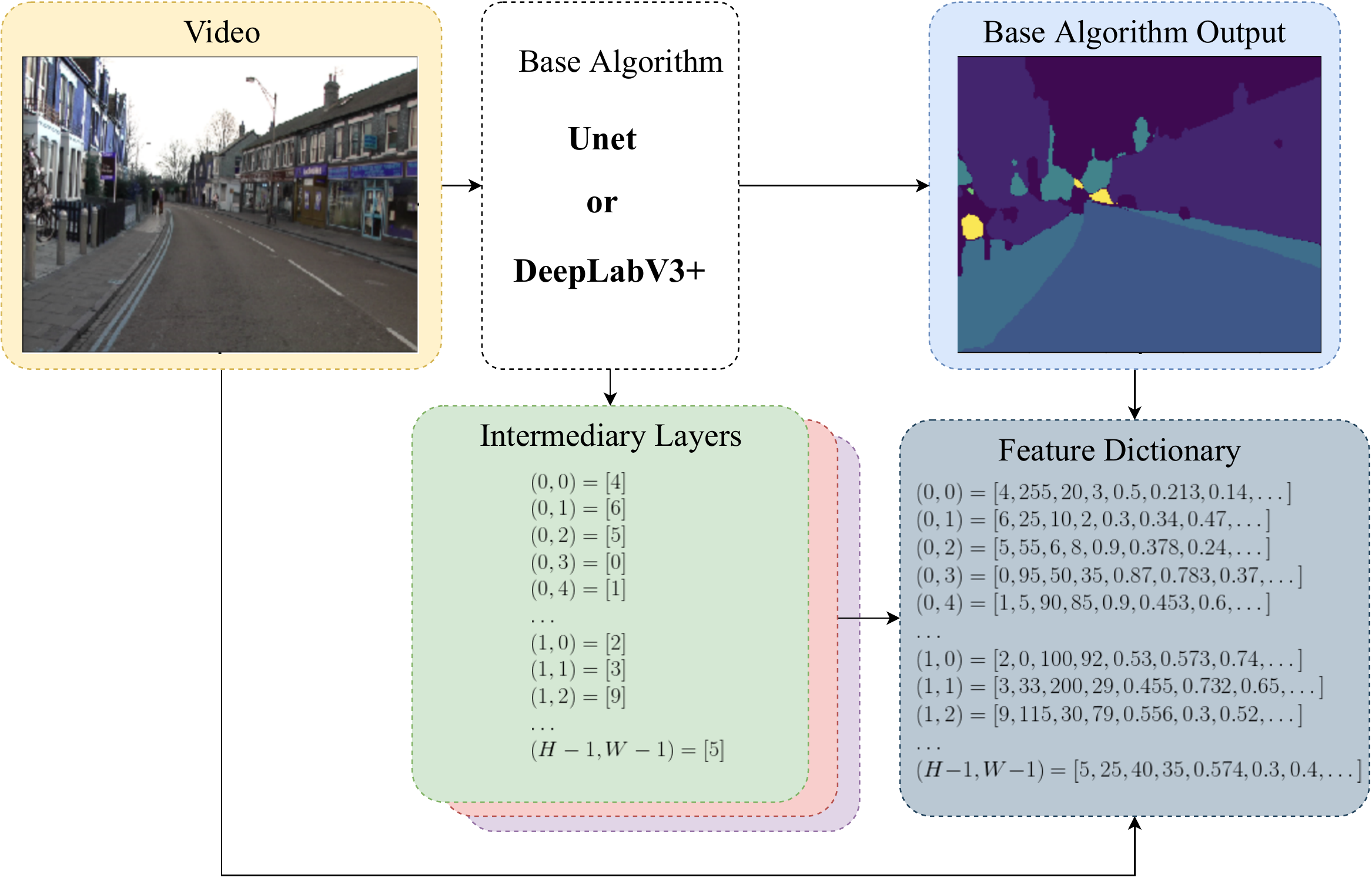} 
        \vspace{0.25cm}
    \caption[Feature extraction pipeline]{Pipeline for feature extraction from the initial frame. Each feature set is concatenated into a single feature dictionary with the pixel coordinates being the keys of the hash map. Each pixel has at least the class predicted by the base algorithm and the associated RGB values. Intermediate layer values are very dependant on the base algorithm.}
    \label{fig:Feature extraction pipeline}
\end{figure}

Features used for each node consist of the intensity values from the three RGB channels of the video and the output segmented image from the base algorithm such as from Unet or DeepLabV3+. All other features consist of the intermediate values. In case of Unet, the final layers provide features and for the DeepLabV3+, the intermediate layers are used as features for input to GCN. Figure \ref{fig:Feature extraction pipeline} illustrates the pipeline for feature extraction.

\subsection{Connecting the Graph}
\label{subsec:Connecting the graph}

Based on the Boolean mask, nodes are connected that are on the border of objects. The rest of the nodes are either isolated or have only incoming connections.

\begin{figure}[!htb]
    \centering
    \includegraphics[width=0.8\linewidth]{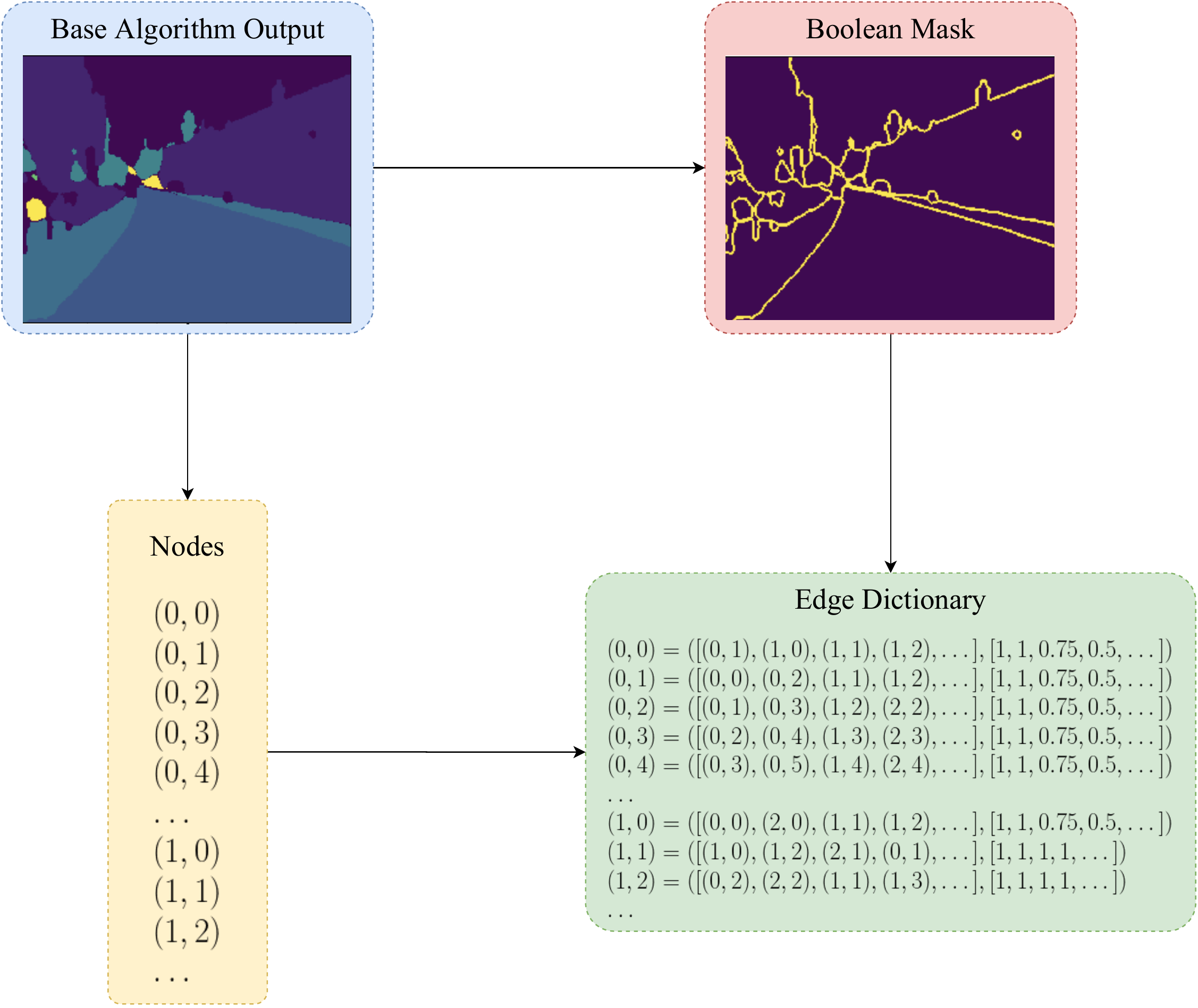} 
        \vspace{0.25cm}
    \caption[Edge generation pipeline]{Pipeline for generating the edges of the graph. The first segmented image provides the size of the image and thus all the potential nodes that need to be processed. The Boolean mask directs which nodes should actually be worked on. If a pixel does not reside on a border then it will remain an isolated node.}
    \label{fig:Edge generation pipeline}
\end{figure}

For each node, the mapping consists of two parts: The actual coordinates of the N closest neighbours denoted by $(x_i, y_i)$ with $i<N$, and the weights that are associated with each of those neighbours denoted by $d_i$ with $i<N$. The illustration of the pipeline for generating the edges is shown in Figure \ref{fig:Edge generation pipeline}.

The weights associated for the edge between two neighbours are completely arbitrary. We chose to use a weight definition that takes into account the distance and intensity differences as inspired by \cite{graphcut}. The resulting weight definition is shown in Equation \ref{eqn:weights}. 
\begin{equation}
    \centering
    d_{n_{12}} = \frac{1}{d_{E_{12}}}\times\exp({-\frac{ \left\lVert I_{n_{1}} - I_{n_{2}}\right\rVert}{ \left\lVert I_{255}\right\rVert}})
     \label{eqn:weights}
\end{equation}
Equation \ref{eqn:weights} represents the weight between two nodes $n_1$ and $n_2$ in the $[0,1]$ range. The inverse of the Euclidean distance between these two nodes denoted as $d_{E_{12}}$. The difference in intensities of the two pixels is taken into account for determining the weight of the edge. This value is normalized and its exponent is taken to penalise two nodes of different values. Considering that several operations have to be performed on each neighbour of every selected node, the overall computational cost is $O(n^k)$ with $n$ being the number of pixels selected, and $k$ the number of closest neighbours we would like to connect to. The Boolean mask assists in directing the resources. If a node is not marked as being on the border, the number of neighbours collapses to 0 in our dictionary. Thus, no computation is performed.

\begin{figure}[!htb]
    \centering
    \includegraphics[width=0.7\linewidth, height=5cm]{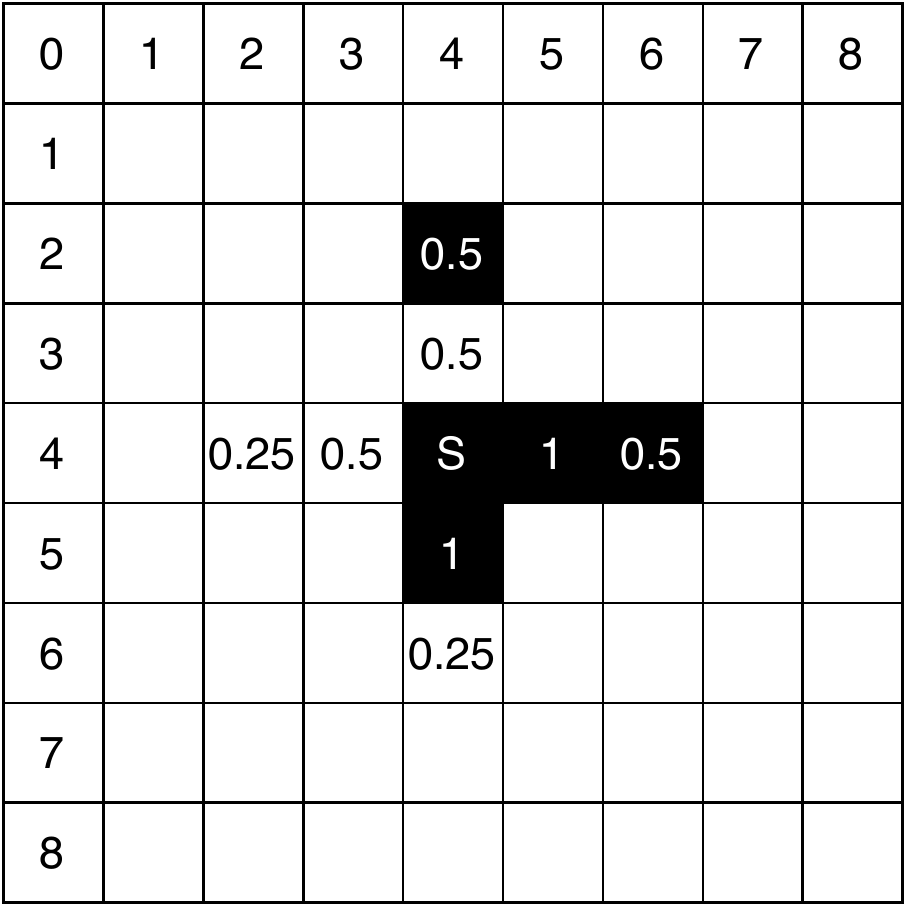}
        \vspace{0.25cm}
    \caption[Our connection weights]{Weights taking into account Euclidean distance and difference in pixel-intensity values.}
    \label{fig:Our connection weights}
\end{figure}

Figure \ref{fig:Our connection weights} illustrates the behaviour of Equation \ref{eqn:weights}. The selected node is denoted as S on the grid. The values in each square represent the value of the weight between S and the node that contains the value. If the colours match and the node is directly adjacent, then the value is maximal. On the other hand, two pixels can be next to each other but with different shades of the same colour. In this case, the value will decrease. We thus encode both the spatial information and also the pixel intensity similarity.

\subsection{Encoding the Graph}
\label{subsec:Encoding the graph}
The input to GCN requires the generation of the associated adjacency and feature matrices. The resulting size of these two matrices depend on the initial size of the frame and on the feature set that was selected. An example is the CamVid dataset which provides frames of 360 by 480 pixels in a video. This creates a square adjacency matrix of 172'800 by 172'800 pixels for each frame. The feature matrix will depend on the selected features and in the case where only intensity values of 3 RGB channels of frames are used as features, the shape will be equal to 172'800 $\times$ 3.

\section{GCN Training}
\label{sec:GCN training}
\begin{figure}[!htb]
    \centering
    \includegraphics[width=1\linewidth, height=4cm]{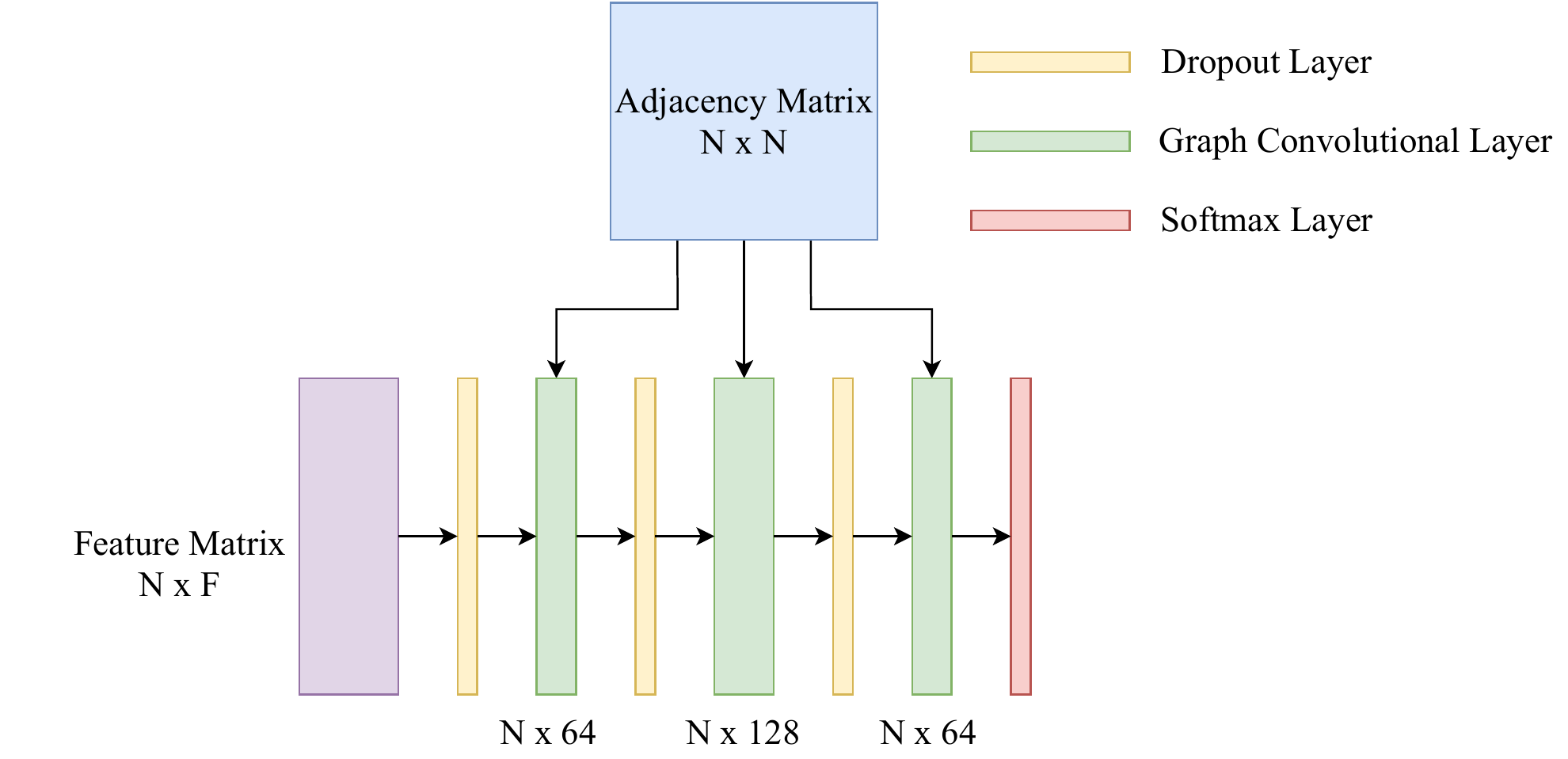} 
    \vspace{0.2cm}
    \caption[Our GCN architecture]{Illustration of our GCN architecture based on \cite{DBLP:journals/corr/KipfW16}.}
    \label{fig:Our GCN architecture}
\end{figure}

After the data has been correctly generated as explained in the previous sections (\ref{sec:Efficient graph generation}-\ref{subsec:Encoding the graph}), the training is carried out. The employed GCN architecture in our Border-SegSGN is a modification on the network given in \cite{DBLP:journals/corr/KipfW16}. The architecture used consists of consecutive graph convolutional layers and dropout layers as shown in Figure \ref{fig:Our GCN architecture}. We experimented with several different versions of network architecture but the Figure \ref{fig:Our GCN architecture} has better performance.

The graph convolutional layers take two input matrices: the adjacency matrix and the feature matrix. The adjacency matrix is considered immutable and thus does not change between the layers. On the other hand, the feature matrix varies with each layer.

\begin{figure}[!htb]
    \centering
    \includegraphics[width=\linewidth, height=7.5cm]{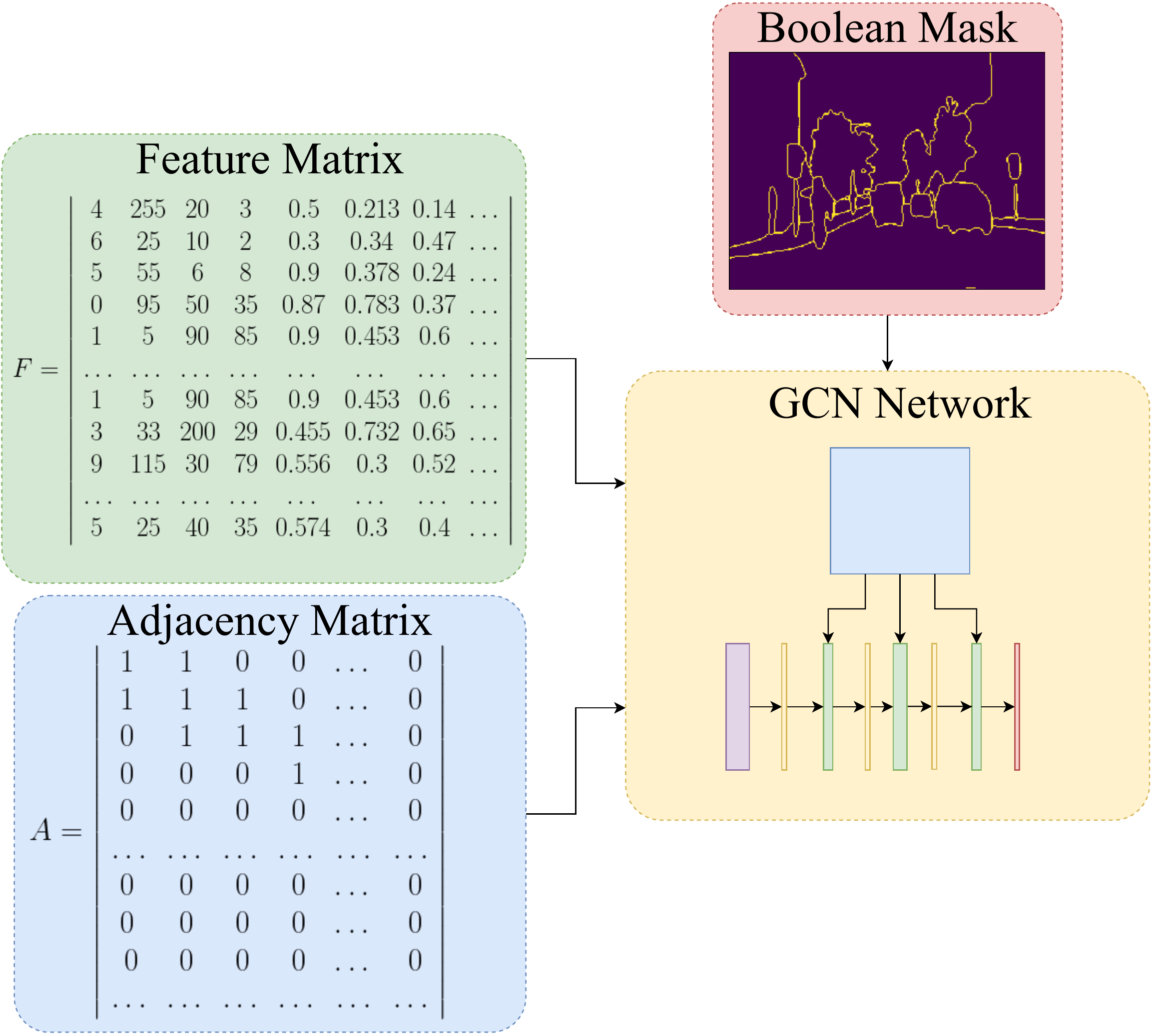} 
    \vspace{0.2cm}
    \caption[GCN training with Boolean mask]{Pipeline representing the inputs for the GCN training in Border-SegGCN.}
    \label{fig:GCN training with boolean mask}
\end{figure}

However, most importantly, we use the Boolean mask created before to direct the GCN training. For the model to fit the correct labels, we pass the mask giving no training relevance on the loss function for the nodes that are not on the border. Figure \ref{fig:GCN training with boolean mask} shows the different inputs to the GCN Network.

\section{GCN Prediction}
\label{sec:GCN prediction}
\begin{figure}[!htb]
    \centering
    \includegraphics[width=\linewidth, height=7.5cm]{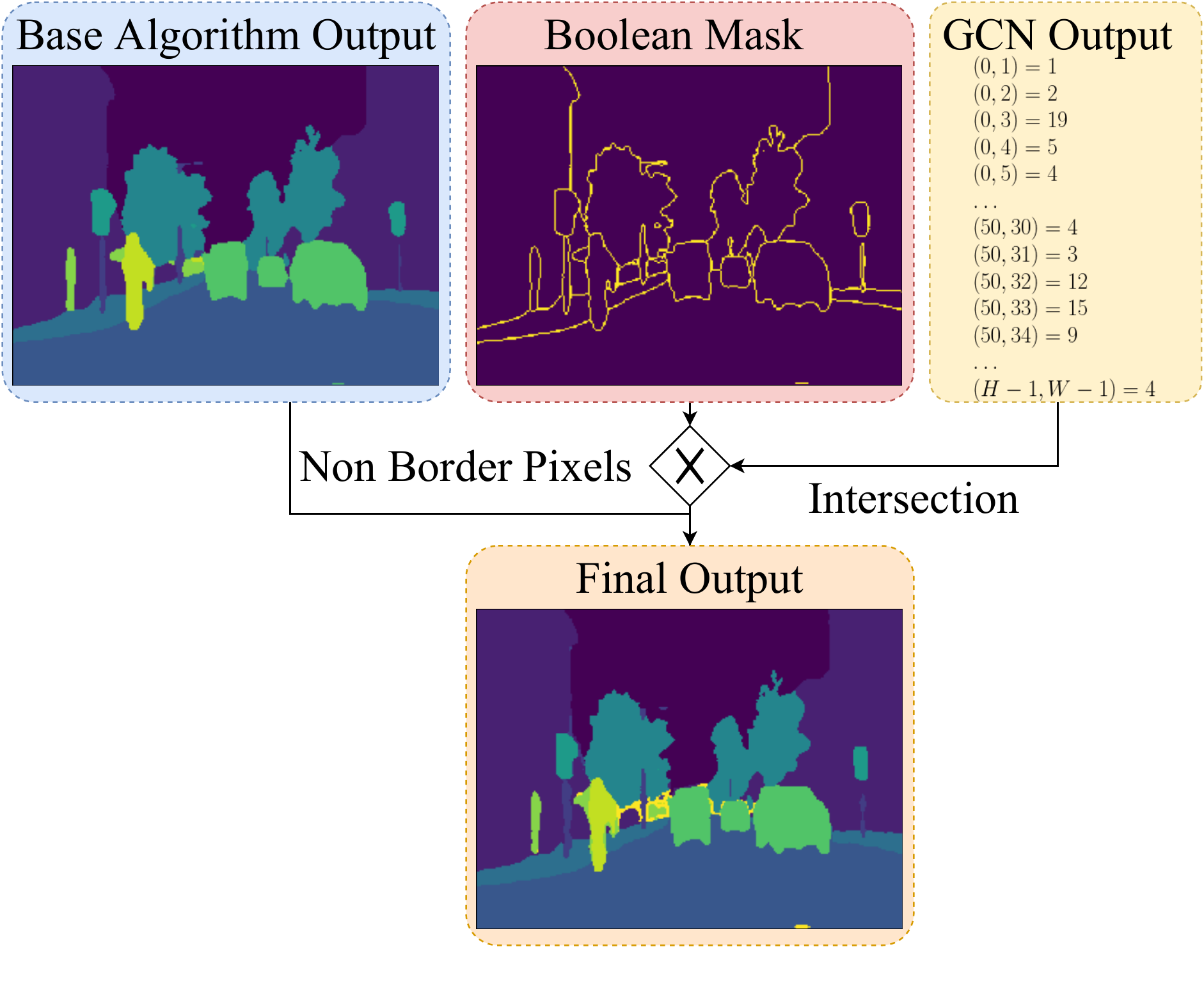} 
    \caption[GCN prediction pipeline]{Pipeline of required inputs to generate final segmented image in Border-SegGCN.}
    \label{fig:GCN predicition pipeline}
\end{figure}

For the prediction as shown in Figure \ref{fig:GCN predicition pipeline}, the GCN takes as input the desired adjacency matrix and feature matrix, but does not require the Boolean mask. We predict the class of every node. Due to the nature of the training, the GCN output is poor on pixels that do not lie on any borders. The solution is to combine the base output of the initial algorithm with the output of the GCN where non-border pixels are assigned to their classes from the output of the base algorithm, while border pixels are assigned to their classes from the GCN output. 

\section{Experiments}
\label{chp:results}
In this section, we verify the effectiveness of our proposed approach for semantic segmentation.

\subsection{Implementation Details}

 We use the Unet implementation of the "Segmentation models" PyTorch library \cite{Yakubovskiy:2019}. The encoder is ResNeXt-50 (32$\times$4d) and the weights of the network are pre-trained from ImageNet. 
For DeepLabV3+, we use the openly available pre-trained model from \cite{semantic_cvpr19} based on\cite{chen2018encoderdecoder}. For both models, we register forward hooks after each layer and retrieve data at each step for every passing frame.
The GCN architecture is modelled based on \cite{DBLP:journals/corr/KipfW16} with slight modifications. It consists of three graph convolutional layers with 64, 128 and 64 channels respectively. A dropout layer is inserted between each graph convolutional layer.
The GCN is built in Keras using the "Spektral" library \footnote{https://spektral.graphneural.network} and an Adam \cite{kingma2014adam} optimiser with a learning rate of 0.001. This approach assumes that each pixel is a sample and this is why it is necessary to set a batch size of $N = H \times W$, where $H$ is the height of frame and $W$ is its width. As a metric for evaluation, we use the mean intersection over union (mIoU) score as defined in \cite{semantic_cvpr19}.

\subsection{Dataset}
\textbf{Camvid} We use the "CamVid" \cite{BrostowFC:PRL2008} dataset for the evaluation of Border-SegGCN. It consists of 367 training, 233 testing, and 101 validation fully annotated frames. Each frame is $360 \times 480$ pixels. The testing frames are from two different video sequences, Seq05VD and 0001TP. We do not perform any kind of augmentation on the data for training. 
We work with the 11 recommended classes out of the 31 available ones to be able to compare with other literature that have followed similar approaches of using 11 classes \cite{semantic_cvpr19}.

\textbf{Carla} \cite{dosovitskiy2017carla} is a semantic segmentation dataset for self-driving cars. t is a simulated dataset and not the real-world images. It is available as open-source. The original dataset has 4550 training images and 449 testing images. We used a subset of this dataset, i.e., 500 for training, 100 for validation, and 300 for testing. We pre-processed this dataset to have the same frame-size as CamVid dataset i.e. $360 \times 480$ pixels. We cropped the image from the center of the image. There are 13 classes for the pixels to be classified on in the dataset.
\subsection{Quantitative Results}

Table \ref{Tab:conclusion} illustrates the mIoU results for the Border-SegGCN using Unet and DeepLabV3+ models as base segmentation networks on CamVid and Carla datasets.

\begin{table}[!htb]
	\centering
		\begin{tabular}{ccc}
			\toprule
			Metric & DeepLabV3+  & Unet\\
			\midrule
			Base mIoU & 81.63 & 79.77 \\
            Our mIoU & \textbf{81.96} & \textbf{80.49}\\
            Max theoretical mIoU & 89.15 & 83.98 \\
            Base mIoU on border & 39.17 & 34.59\\
            Our mIoU on border & \textbf{40.92} & \textbf{45.67}\\
      	\bottomrule
      	
        \end{tabular}
                 \vspace*{0.3cm}
       
	\normalsize
    \caption[Results considering context]{Quantitative mIoU values achieved using Border-SegGCN with UNet and DeepLabV3+. Note that the theoretical maximum is computed by replacing all labels for pixels on the border with their ground truth.}
	\label{Tab:conclusion}
\end{table}

DeepLabV3+ has been shown to outperform the vanilla Unet. So, it has more border pixels than the Unet model. Hence, DeepLabV3+ model has a higher theoretical maximum. The relative improvement on the Unet is much larger than that on the DeepLabV3+ model because of a lower upper limit. Thus, having less problematic borders to re-classify.
The best performing mIoU on the CamVid dataset using our Border-SegGCN achieved 81.96. In \cite{semantic_cvpr19}, they obtained a mIoU of 81.7. Using their code, we managed to reproduce 81.63. We were able to register a real improvement on this as baseline by 0.404\%. Furthermore, it is important to consider that the border pixels only constitute between 10-20\% of the entire image. This effectively creates an upper limit on the performance that the Border-SegGCN can not exceed. 

We do not train Carla dataset for two epochs with Unet because if trained till the best mIoU then most of the border pixels in the image are correctly classified for this dataset. We want to show with our algorithm and experiments that if there are wrong pixels classified along the border with base algorithm then BorderSegGCN helps to rectify those pixel classification on borders. We use same parameter setting for training with Carla dataset as used for with Camvid dataset when Unet is used as baseline.

We performed an ablation study on different parameters such as border thickness, number of edges for each node, number of input features from base network to GCN, effect of using different base network, etc. on CamVid dataset. The results are sumarized in Table \ref{Tab:ablation}

\subsubsection{Border Thickness:} 
\begin{figure}[!htb]
    \centering
    \includegraphics[width=0.31\linewidth]{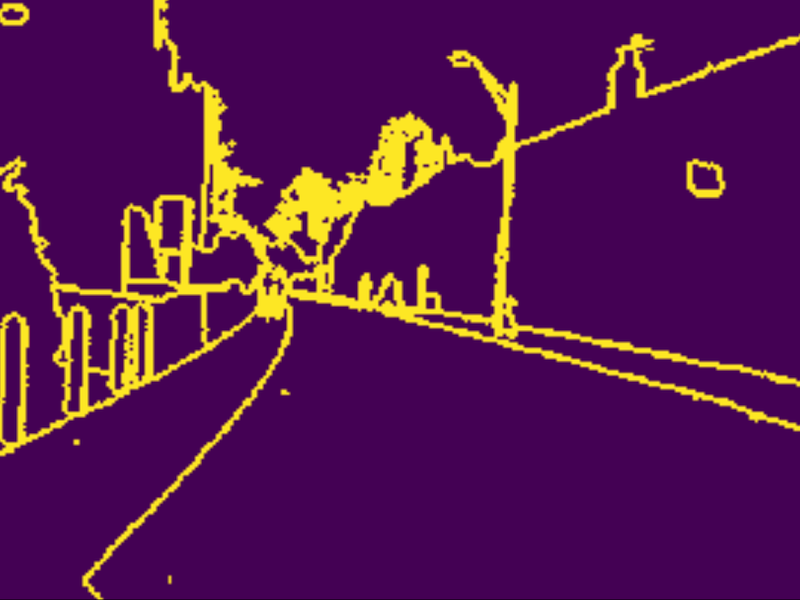}
    \includegraphics[width=0.31\linewidth]{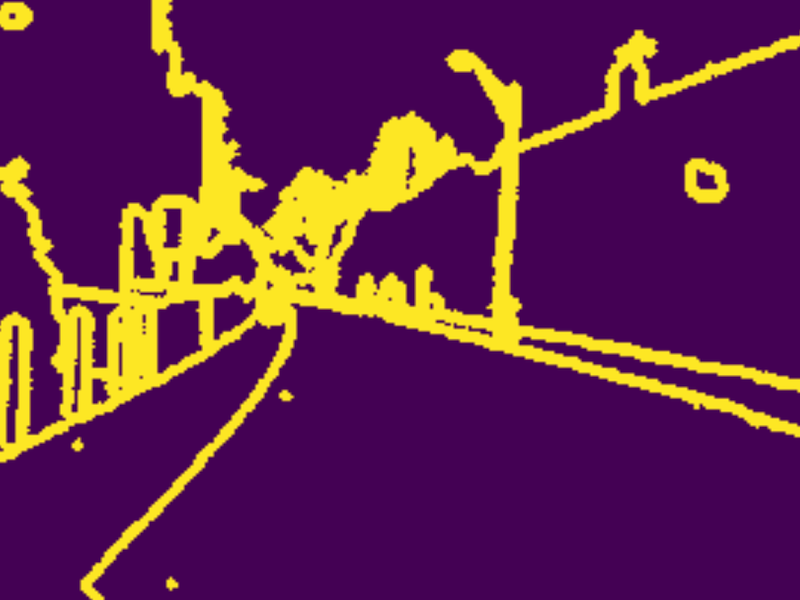}
    \includegraphics[width=0.305\linewidth]{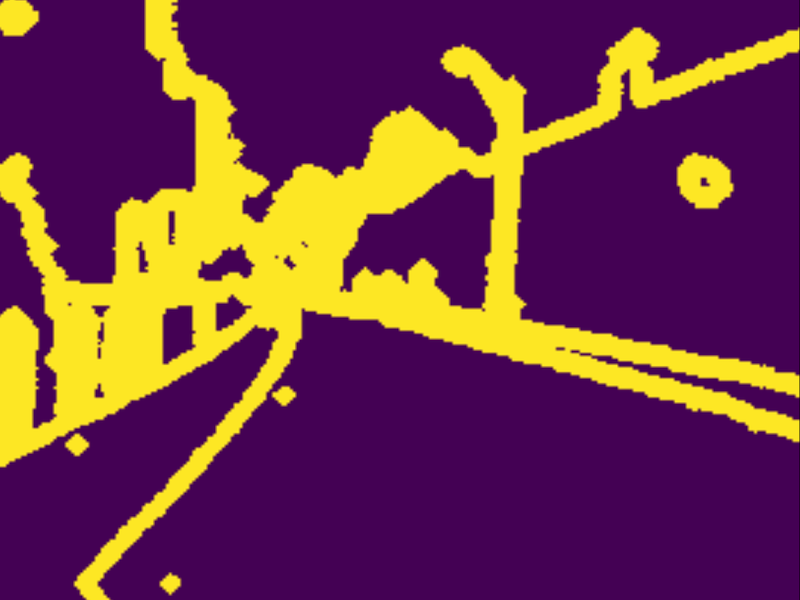} 
    \vspace{0.2cm}
    \caption[Border thickness]{Left to right:  1, 2, 3 border pixel are selected respectively. Frames taken from CamVid dataset.}
    \label{fig:Border thickness}
\end{figure}

\begin{figure}[!htb]
    \centering
     \subfloat[Influence of border thickness.\label{fig:PixelThickness}]{%
        \includegraphics[width=2.3in]{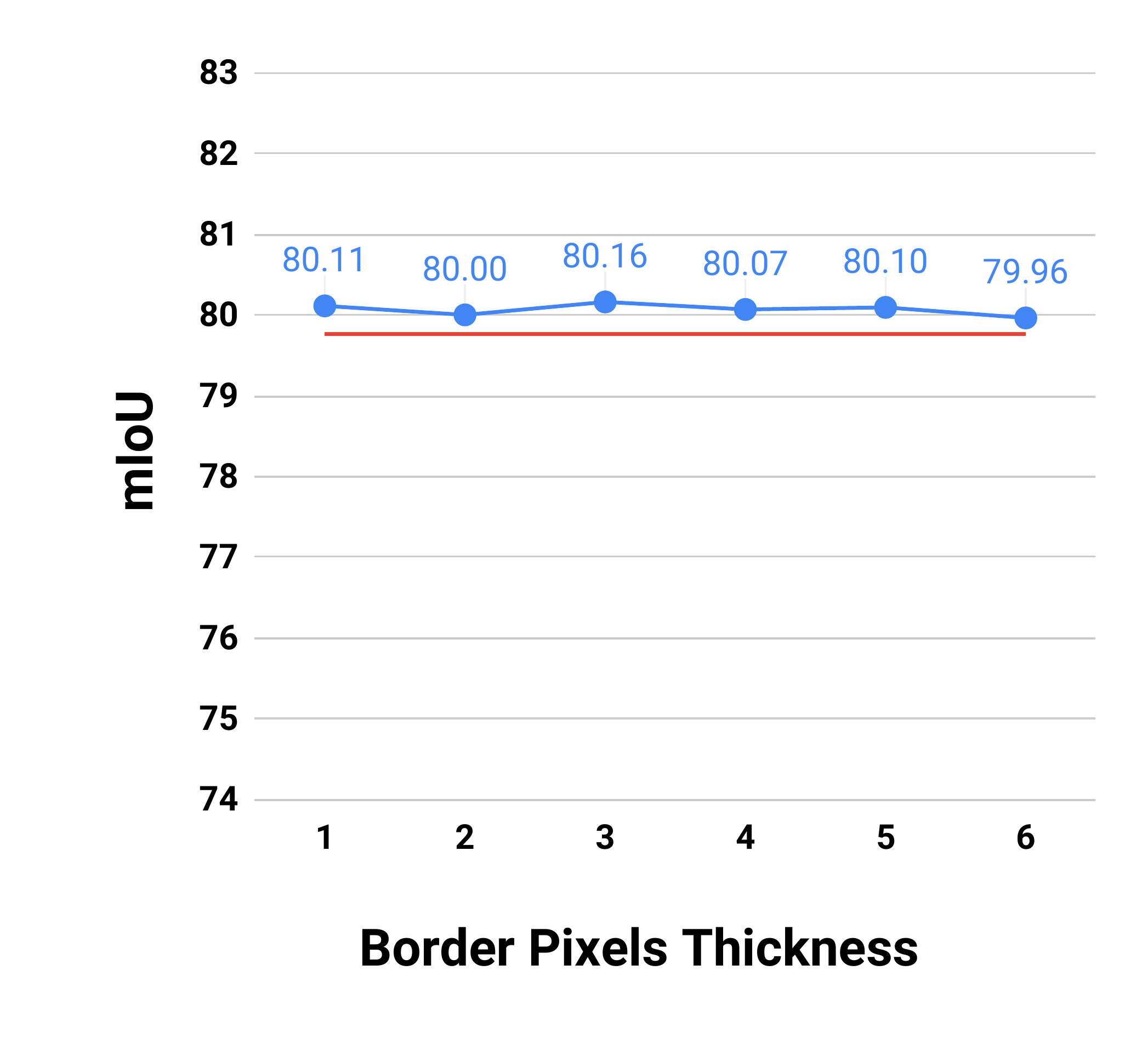}%
      }
      \hfill
     
    \subfloat[Influence of number of connections.\label{fig:NumCon}]{%
            \includegraphics[width=2.3in]{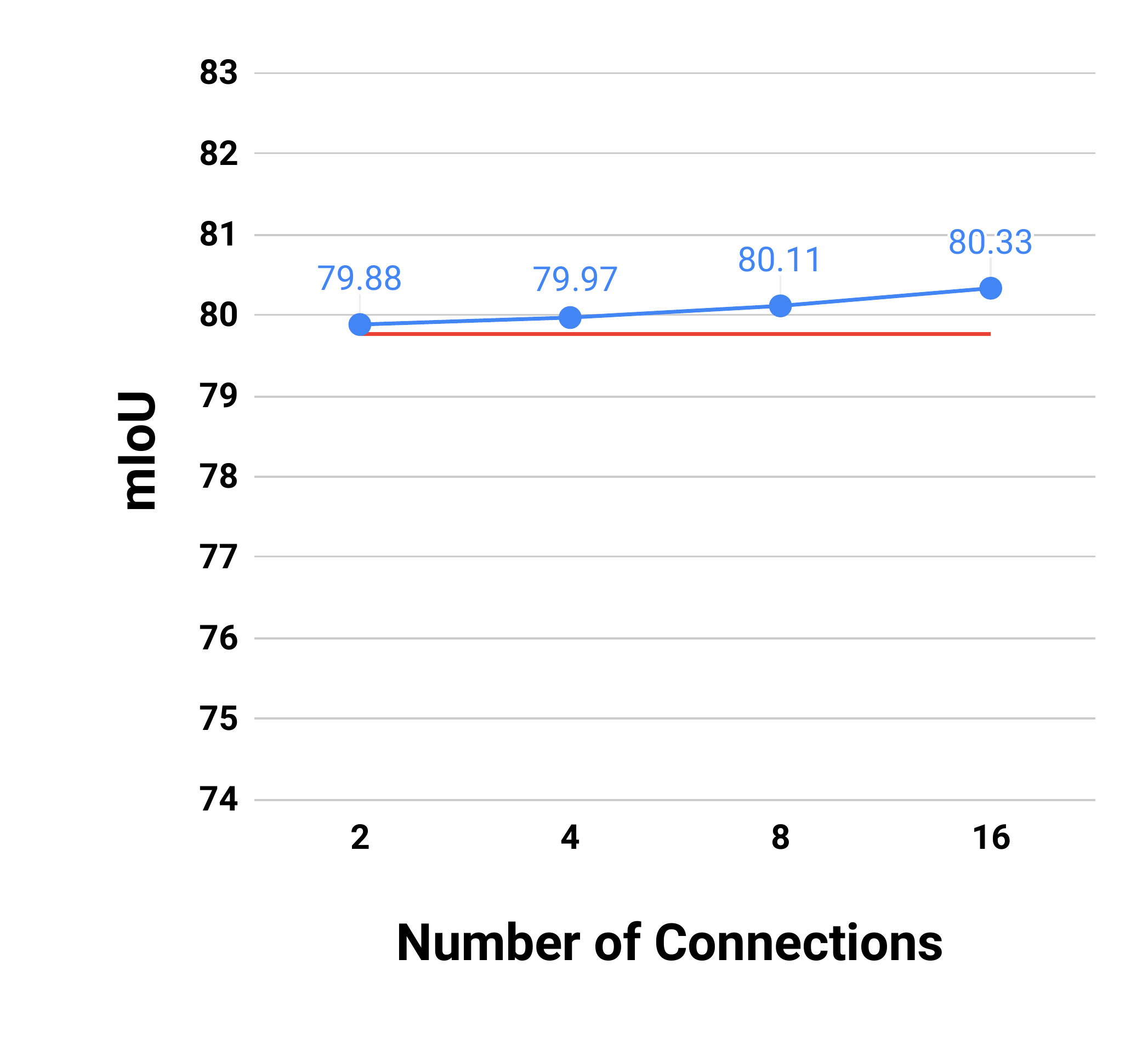}%
            }

      \vfill
    \includegraphics[width=1.2in]{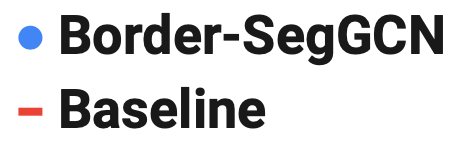}%
    \caption[Influence of border thickness]{Influence of graph attributes on mIoU.}
    \label{fig:GraphVar}
\end{figure}

Figure \ref{fig:Border thickness} shows the camvid dataset frames with different border pixel thickness. Figure \ref{fig:PixelThickness} shows that except certain outliers, the overall trend is that as the amount of border pixels increases, the mIoU decreases. This can be due to the spatial intrinsic characteristics that border pixels have, which allows the GCN to better fit to them. 

\begin{table*}[!htb]
	\centering
	\begin{tabular}{cc|cc|cc|cc|ll}

			\toprule

        \# connections & mIoU & Border Thickness & mIoU & \# features & mIoU  & Dropout Rate &  mIoU & Regularization  & 	mIoU\\
	\hline
			   2 & 79.88 & 1 & 80.11 &  11 & 78.37  & 0 &  80.20 & 1.00e-01  & 	74.79\\
               4 & 79.97 & 2& 80.00 & 15 & 80.45  & 0.0001 & 80.39 &  1.00e-04 & 80.12 \\
               8 & 80.11 & 3& 80.16 & 256 & 81.12 & 0.001 &  80.11 & 1.00e-08   &80.24\\
               16 & 80.33 & 4 &  80.07 & 271 & 80.90 & 0.1 & 80.09 & 1.00e-11 & 80.14 \\
               - & - & 5&  80.10 & 319 & 80.99 & 0.5 &  80.33 & 1.00e-13 & 80.22\\
               - & - & 6 & 79.96 & 510 & 81.59&  0.9 & 79.39 & - & -\\
               - & - & - & - & 1039 & 81.73 &   - & -&   - & -\\
               - & - & - & - & 1999 & 81.92 &  - & - &   - & -\\

      	\bottomrule
      	
        \end{tabular}
      	 \vspace*{0.1cm}
       
	\normalsize
	      	 \vspace*{0.2cm}

    \caption[Results considering context]{Ablation study with quantitative mIoU values achieved using Border-SegGCN with DeepLabV3+ for Camvid dataset using different number of connections in the graph, number of border pixel thickness, number of features, dropout rate, and regularization}
	\label{Tab:ablation}
\end{table*}

\subsubsection{Number of Edges:} As the number of edges is variable, we see how the number of connections each node has influences the results of improving the segmentation task (with the help of improving the border pixel classification). Figure \ref{fig:NumCon} shows that the results steadily increase as the number of edges increases because the more edges a singular node has, the more spatial information it carries about its neighbours. It is also worth mentioning that increasing indefinitely the number of edges has two adverse effects. First, it becomes computationally more expensive to generate the graph as more information is stored per node. But second, and more importantly, the impact of each edge decreases as the number of edges increases. 

\subsubsection{Number of Features:} 
\begin{figure}[!htb]
\centering
    \subfloat[Border-SegGCN with Unet\label{fig:Unet feat1}]{%
            \includegraphics[width=2.3in]{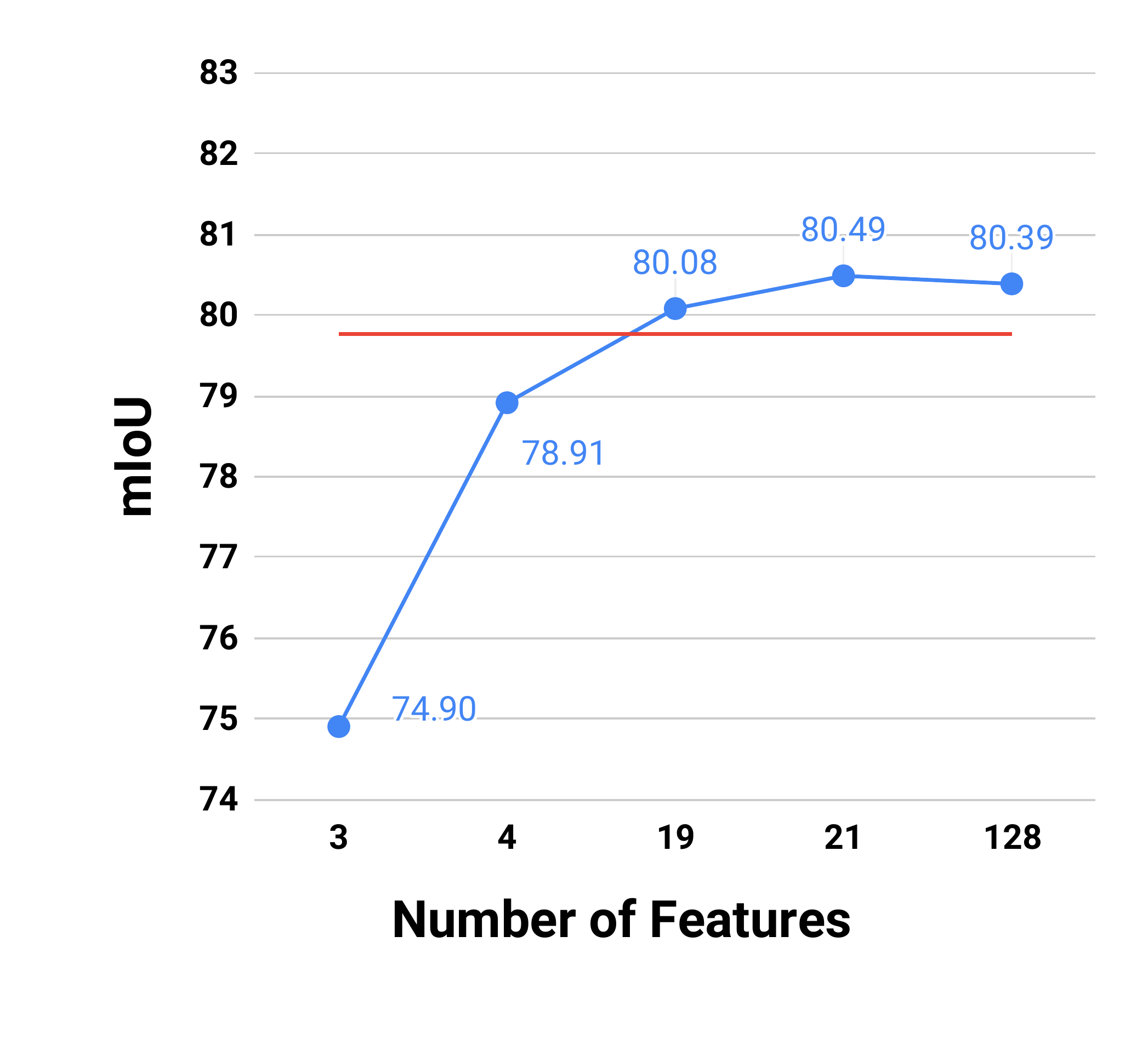}%
            }
            \hfill
      \subfloat[Border-SegGCN with DeepLabV3+\label{fig:DeepLab feat1}]{%
        \includegraphics[width=2.3in]{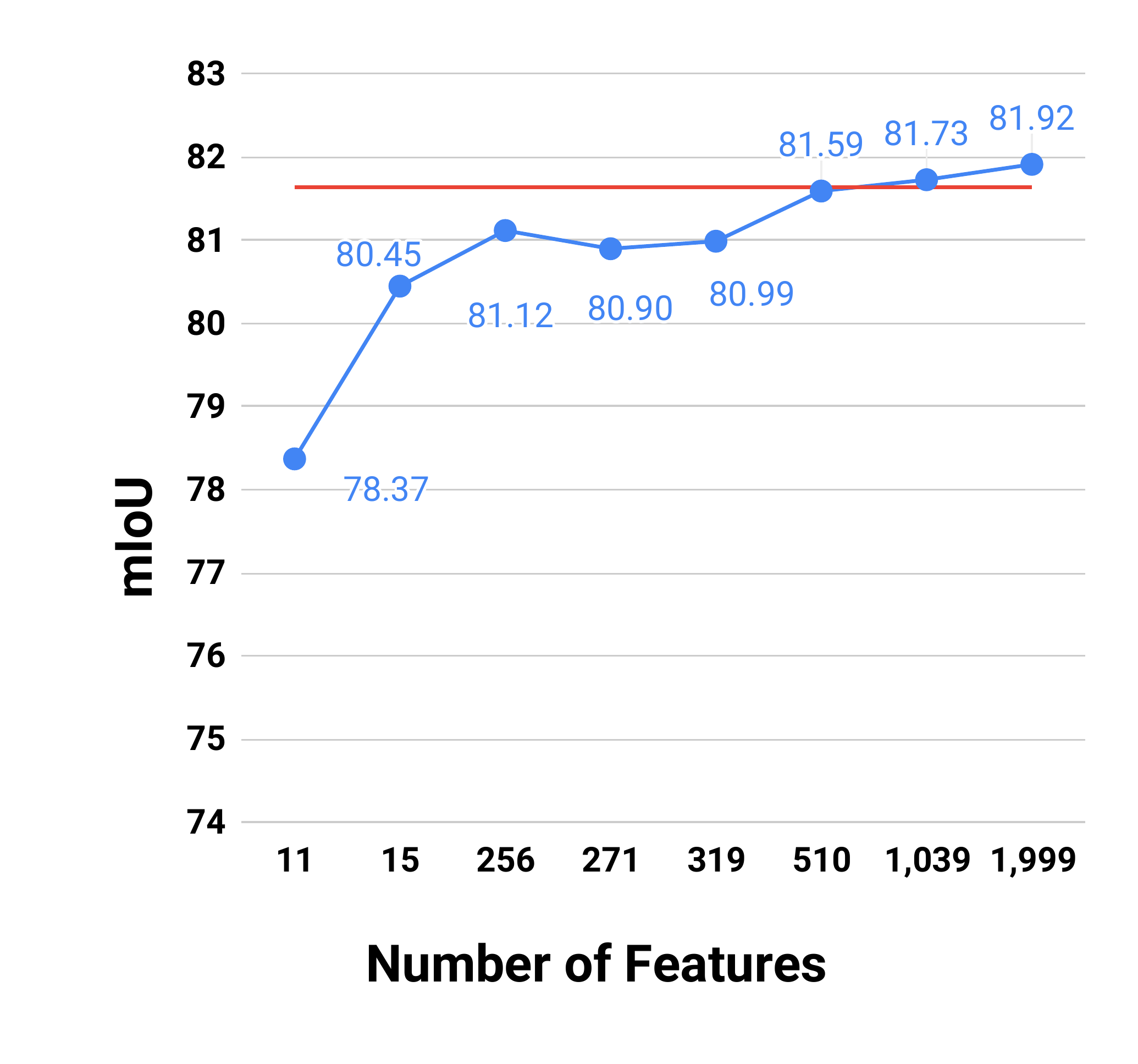}%
     
      }
      
      \vfill
        \includegraphics[width=1.2in]{figures/legend.png}%
 \caption[Influence of number of features]{Influence of number of features on the mIoU.}
    \label{fig:FeatSeg}
\end{figure}

The most important consideration within the graph creation process is the feature selection step. The features have to represent enough information to the GCN for it to predict the correct pixel class. Figure \ref{fig:FeatSeg} represents the correlation between the mean IoU and the number of feature channels.

Each step represents a different set of features, representing a combination of different feature sets. Unsurprisingly, the best results are obtained by combining the base algorithm's output, the intensity values of the frames, and a combination of intermediary layers.

\begin{table}[!htb]
	\footnotesize
	\centering
		\begin{tabular}{cc}
			\toprule
			Number of features & Constituent features\\
			\midrule
			3 &  'I' \\
            4 & 'base' + 'I'\\
            19 & 'RGB' + 'd5' \\
            21 & 'base' + 'I' + 'seg' + 'd5' \\
            128 & 'base' + 'I' + 'seg' + 'd3' + 'd4' + 'd5'\\
    
			\bottomrule
        \end{tabular}
        \vspace*{0.3cm}
        	\normalsize
    \caption[Unet number of feature mapping]{Decomposition of the features that were used as input to GCN when Unet is a base network in Border-SegGCN. Notations: I-intensity values of the RGB channels, base-output segmented image from base network, d3-d5-last 3 decoder layer of Unet.}
	\label{tab:imprwdpatterns3}
\end{table}

\begin{table}[!htb]
	\footnotesize
	\centering
		\begin{tabular}{cc}
			\toprule
			Number of features & Constituent features\\
			\midrule
			11 &  'I' \\
            15 & 'I' + 'base' \\
            256 & 'mba' \\
            271 & 'base'+'I'+ '$final_{6}$'+'$final_{3}$'\\
            319 & 'base'+'I'+ 'seg'+'mba'\\
            510 & 'I'+ '$final_{6}$'+'m1+'m2'+'seg'+'mba'\\
            1039 & 'I'+ '$final_{6}$'+'m5'\\
            1999 & 'base'+ 'I'+ '$final_{6}$' +'m1'+'m2'+'m3'+'m4'+'m5' \\
    
			\bottomrule
        \end{tabular}
                \vspace*{0.3cm}
	\normalsize
    \caption[Unet number of feature mapping]{Decomposition of the features  that were used as input to GCN when DeepLabV3+ is a base network in Border-SegGCN. Notations: I-intensity values of the RGB channels, base-output segmented image from base network,  '$final_{1-6}$'-Segmentation heads sublayers 1 to 6, m1-m7- Encoder layers 1 to 7, 'mba'-Atrous spatial pyramid pooling output}
	\label{tab:imprwdpatterns}
\end{table}

\subsubsection{GCN Dropout Optimisation:}
\begin{figure}[!htb]
\centering
    \subfloat[Dropout effect on mIoU\label{fig:Dropout}]{%
            \includegraphics[width=2.3in]{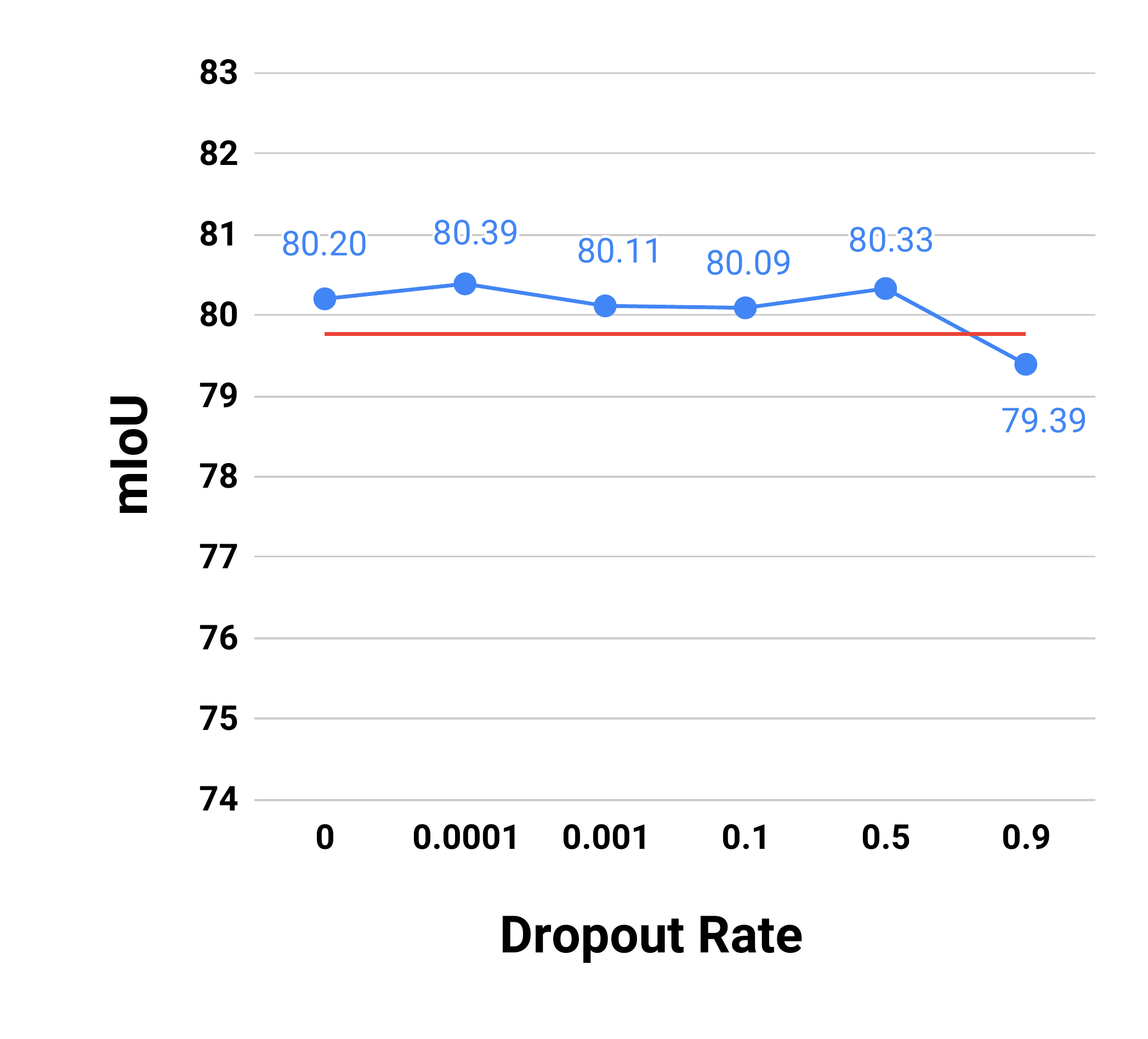}%
            }
            \hfill
      \subfloat[Regularisation effect on mIoU\label{fig:Regularisation}]{%
        \includegraphics[width=2.3in]{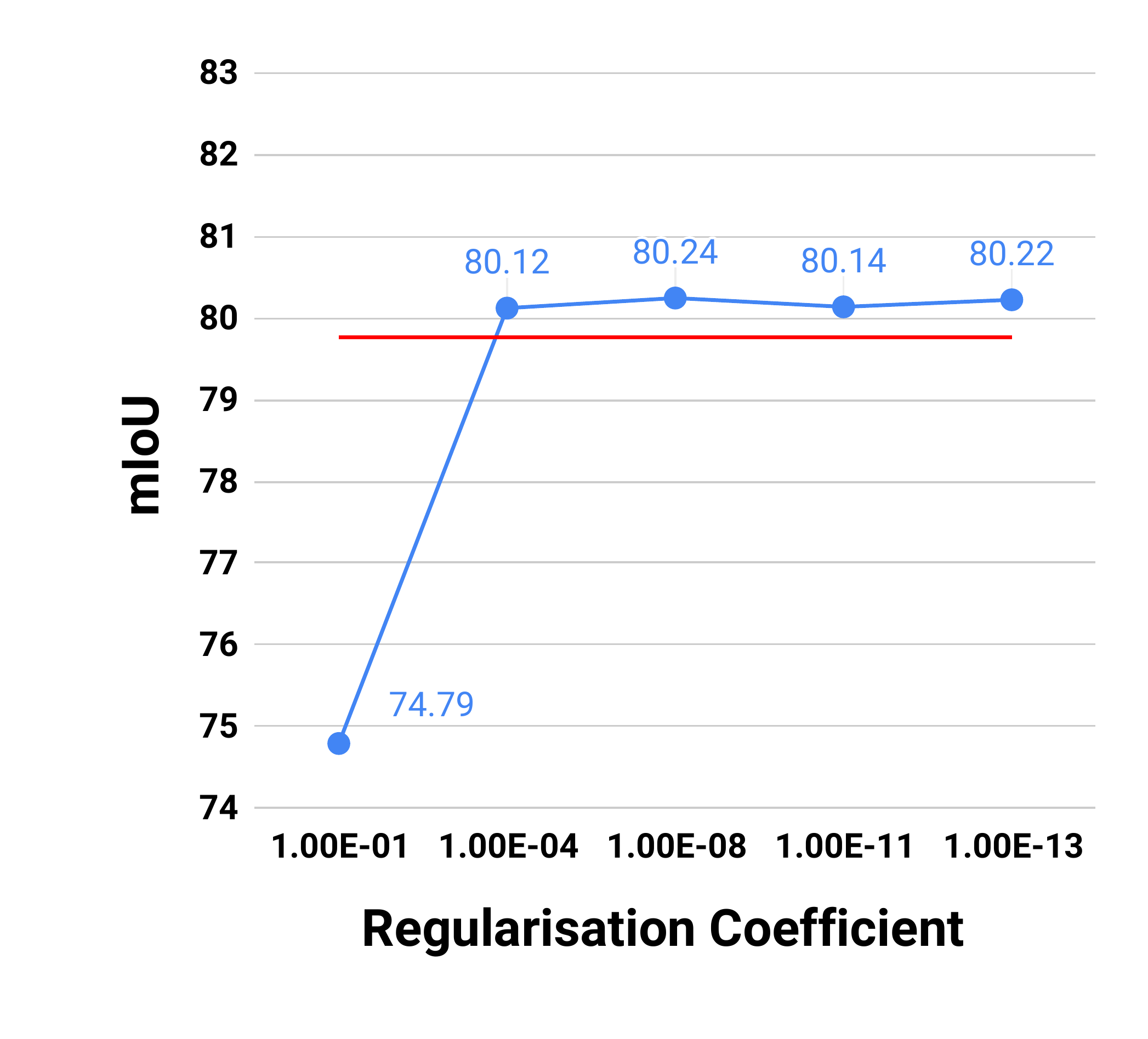}%
      }
      
      \vfill
    \includegraphics[width=1.2in]{figures/legend.png}%
    \vspace*{0.2cm}
    \caption[Dropout and regularisation]{Influence of dropout and regularisation on Border-SegGCN.}
    \label{fig:DropReg}
\end{figure}

  We used a modification of the architecture given in \cite{DBLP:journals/corr/KipfW16} that showed an improvement when the number of layers is varied from the original network. We show the dropout effects on the mean IoU score in Figure \ref{fig:Dropout}.

\subsubsection{GCN Regularisation Optimisation:}
Figure \ref{fig:Regularisation} shows the mIoU improvement as the regularisation coefficient decreases. This concludes that the GCN is not prone to overfitting and thus can generalise the results. 

\subsubsection{Different Base Model: DeepLabV3+:}
We use the base network Unet in the previous studies. Recent research shows that there have been architectures that have improved performance over Unet such as the DeepLabV3+ \cite{chen2018encoderdecoder}.

We observe the same trend with the Unet experiments. When increasing the number of feature channels, mIoU increases as well. However, the amount of features is much higher and requires much more computational resources with this network. Figure \ref{fig:DeepLab feat1} suggests that we are capable of improving an output from different models with our architecture pipeline.

\subsection{Time Complexity}
Total time required for each frame in the video is around 19 seconds. The average time for DeepLabV3+ is 0.08sec per frame. So, Border-SegGCN has a drawback of latency. We are introducing this technique as a stepping stone so that the research can also be focused on GCNs for improving semantic segmentation using improved techniques.

\subsection{Image Size}
This approach requires adjacency matrix has squared dimension of the given image. So, due to available GPU resources, it is not possible to do experiments on various video segmentation datasets having large frame size.

\subsection{Qualitative Results}
\label{sec:Output}
\begin{figure}[!htb]
    \centering
     \subfloat[\label{fig:Original}]{%
    \includegraphics[width=0.22\linewidth, height=2cm]{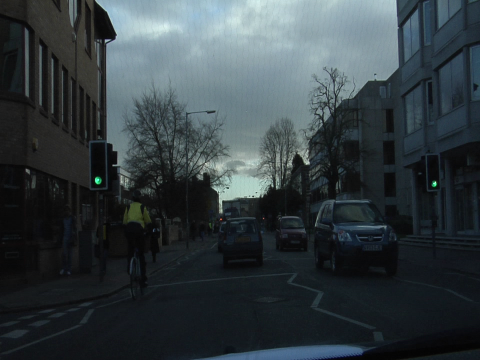}
    }
      \subfloat[\label{fig:DeepLabV3+ baseline}]{%
    \includegraphics[width=0.22\linewidth,height=2cm]{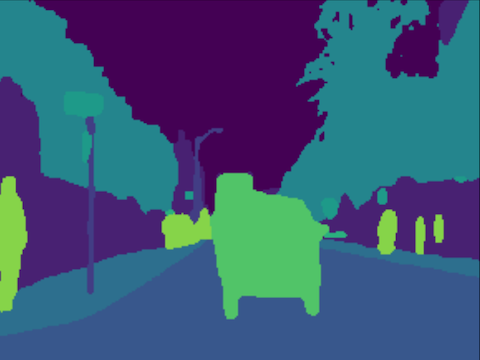}
    }
     \subfloat[\label{fig:Border-SegGCN with DeepLabV3+ good}]{%
    \includegraphics[width=0.22\linewidth,height=2cm]{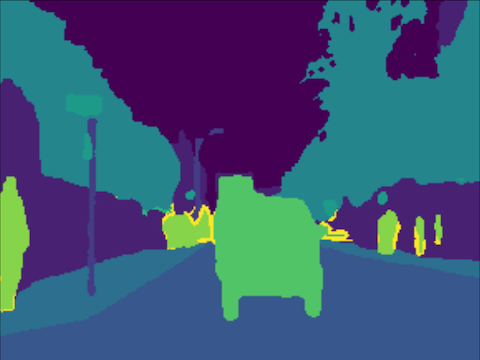}
    }
     \subfloat[\label{fig:Border-SegGCN with DeepLabV3+ bad}]{%
    \includegraphics[width=0.22\linewidth,height=2cm]{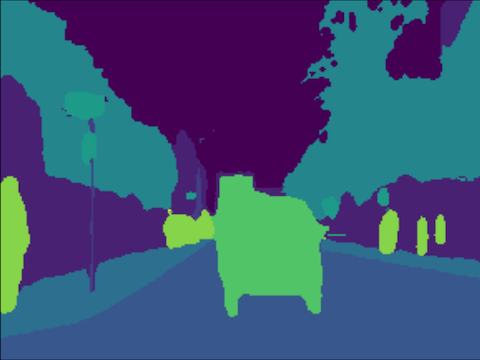}
    }\hfill
     \subfloat[\label{fig:Ground truth}]{%
    \includegraphics[width=0.22\linewidth,height=2cm]{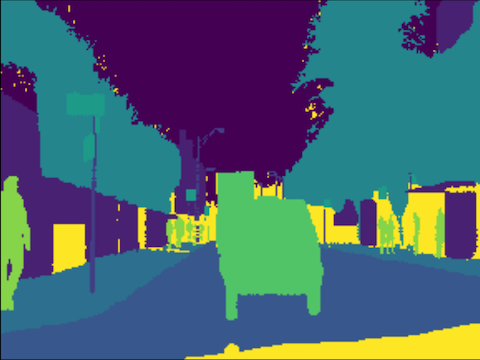}
    }
     \subfloat[\label{fig:Unet baseline}]{%
    \includegraphics[width=0.22\linewidth,height=2cm]{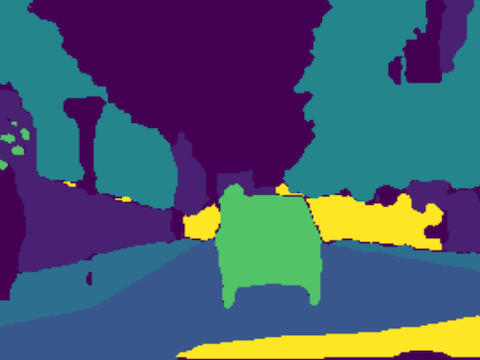}
    }
     \subfloat[\label{fig:Border-SegGCN with Unet good}]{%
    \includegraphics[width=0.22\linewidth,height=2cm]{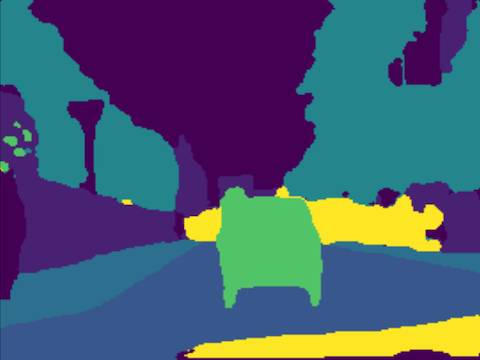}
    }
     \subfloat[\label{fig:Border-SegGCN with Unet bad}]{%
    \includegraphics[width=0.22\linewidth,height=2cm]{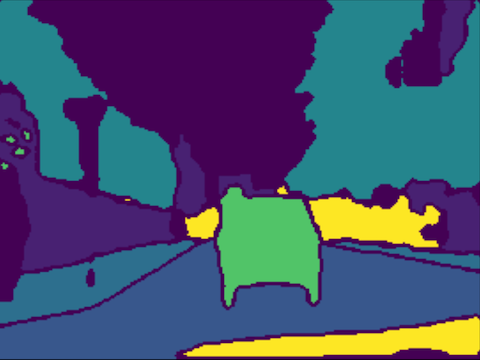}
    }
    \vspace{0.2cm}
    \caption[Output]{(a) Original. (b) DeepLabV3+ baseline. (c) Best Border-SegGCN with DeepLabV3+. (d) Poor Border-SegGCN with DeepLabV3+.
    (e) Ground truth. (f) Unet baseline. (g) Best Border-SegGCN with Unet. (h) Poor Border-SegGCN with Unet.}
    \label{fig:Output}
\end{figure}

Figure \ref{fig:Output} shows qualitative examples of the output generated by our pipeline. The output segmented frames from using both, the Unet and DeepLabV3+, as base network  are illustrated with their best input feature set compared to an under-performing input feature set.

 We noticed from Figure \ref{fig:Output} that 
 \begin{itemize}
  \item  In the case of Unet as shown in Figure \ref{fig:Unet baseline}, when the model is under-performing due to use of wrong feature set input to the GCN, the model defaults to the statistical most likely category \ref{fig:Border-SegGCN with Unet bad}. 
  \item DeepLabV3+ has difficulties with objects that appear smaller in the frame \ref{fig:DeepLabV3+ baseline}, 
  \item Our model is capable of correcting the DeepLabV3+ predictions on the outline of those objects \ref{fig:Border-SegGCN with DeepLabV3+ good}. 
  \item The outlines using Border-SegGCN in \ref{fig:Border-SegGCN with Unet good} are better defined as compared to \ref{fig:Unet baseline}.
 \end{itemize}
\section{Conclusion}
\label{sec:Conclusion}

Our proposed model “Border-SegGCN” employs the base segmentation network such as Unet and DeepLabV3+ along with the GCN. Border-SegGCN is used to refine the object boundaries predictions. The most important task for improving the prediction performance of the base algorithm is the feature selection to be used as input to GCN. The variation due to features is the largest contributing factor in both baseline architectures, i.e., Unet and DeepLabV3+. Our experiments showed that Border-SegGCN is agnostic to the choice of baseline model. The more spatial information a node has on its neighbours, expressed by the number of edges, the better is the performance of the GCN. Finally, our experimental results show that the proposed model gives a new state-of-the-art performance on CamVid dataset using DeepLabV3+ as baseline network. It also improved the results of Unet baseline for both Camvid and Carla datasets. In future work, we will work on tackling the drawbacks of this technique, i.e. improving latency, and reducing computation complexity.

{\small
\bibliographystyle{ieee_fullname}
\bibliography{egbib}
}

\end{document}